\documentclass[letterpaper]{article}
\usepackage[preprint]{arxiv2027}
\usepackage[hyphens]{url}
\usepackage{graphicx}
\usepackage{natbib}
\usepackage{caption}
\usepackage{amsmath,amssymb}
\usepackage{algorithm}
\usepackage{algorithmic}
\usepackage{placeins}
\usepackage{booktabs}
\usepackage[hidelinks]{hyperref}
\title{ReDiPPO: Reference-Guided Value Calibration\\
and Discrepancy-Aware Token Reweighting for Mathematical Reasoning}
\author{
Zhenrong Zhang\textsuperscript{1}\equalcontrib,
Fei Wu\textsuperscript{2}\equalcontrib,
Jun Du\textsuperscript{2},
Jianshu Zhang\textsuperscript{1}\corresponding,
Si Wei\textsuperscript{1}
}
\affiliations{
\textsuperscript{1}IFLYTEK Research\\
\textsuperscript{2}University of Science and Technology of China
}

\begin{document}
\maketitle

\begin{abstract}
Reinforcement learning has emerged as an effective paradigm for enhancing the mathematical reasoning capabilities of large language models. Among existing policy optimization methods, Proximal Policy Optimization (PPO) remains particularly appealing because its learned critic can, in principle, provide token-level credit assignment. However, in mathematical reasoning tasks characterized by long reasoning horizons and sparse outcome rewards, reliable token-level credit assignment remains challenging. The standard critic often fails to accurately evaluate intermediate reasoning states, resulting in noisy advantage estimates and suboptimal policy updates.
In this paper, we propose \textbf{ReDiPPO}, a \textbf{Re}ference-guided and \textbf{Di}screpancy-aware PPO framework for mathematical reasoning. ReDiPPO introduces a reference-guided critic that uses reference answers as training-time privileged signals to provide more accurate value estimation. Meanwhile, it retains a standard critic and quantifies the token-level reference-standard discrepancy between the standard value estimate and the reference-guided value estimate. This discrepancy serves as an indicator of difficult reasoning states and is used to reweight the corresponding token-level advantages during PPO optimization.
Extensive experiments on diverse mathematical reasoning benchmarks demonstrate that ReDiPPO improves value-estimation accuracy and consistently outperforms strong policy optimization baselines, including PPO, DAPO, and GSPO, in final reasoning performance. Our code is available on \href{https://github.com/cii030/ReDiPPO}{GitHub}.
\end{abstract}

\section{Introduction}
\label{sec:introduction}

\begin{figure}[t]
    \centering
    \includegraphics[width=\columnwidth]{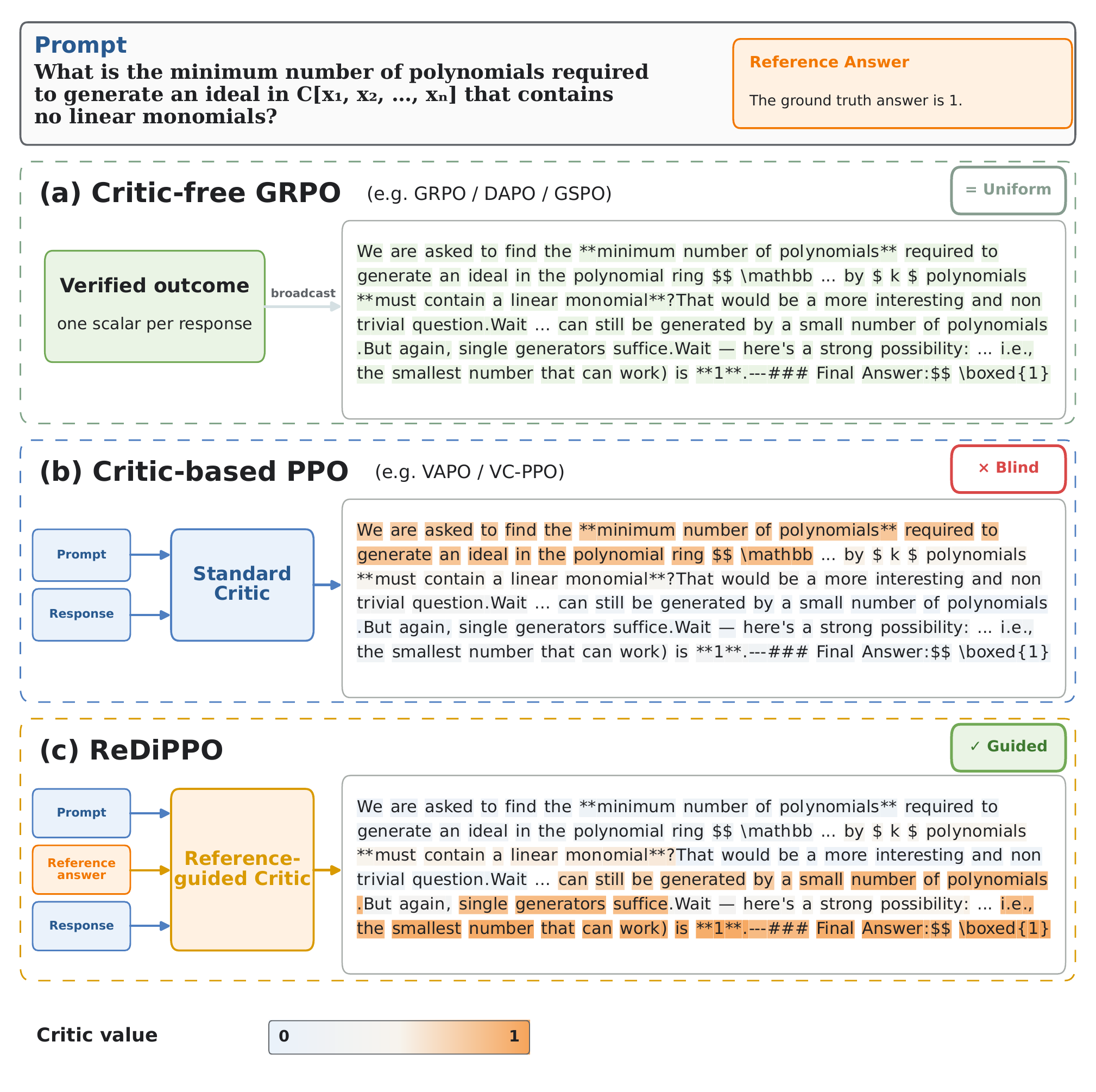}
    \captionsetup{skip=2pt}
    \caption{Qualitative comparison of credit assignment in RLVR. (a) Critic-free GRPO variants broadcast a uniform response-level signal. (b) Critic-based PPO uses a standard critic without access to the reference answer. (c) ReDiPPO conditions its critic on the reference answer for more reliable token-level credit assignment.}
    \label{fig:intro_credit_assignment}
    \vspace{-18pt}
\end{figure}

Reinforcement learning with verifiable rewards (RLVR) has become a central paradigm for improving the mathematical reasoning ability of large language models. In this setting, a model generates a long-horizon reasoning trajectory for a mathematical prompt, and an automatic verifier assigns an outcome reward by comparing the final answer with a reference answer. Recent systems have shown that such outcome-supervised reinforcement learning can substantially improve reasoning performance~\citep{shaoDeepSeekMathPushingLimits2024,guoDeepSeekR1IncentivizesReasoning2025}. Figure~\ref{fig:intro_credit_assignment} summarizes three representative credit assignment signals in RLVR. As shown in Figure~\ref{fig:intro_credit_assignment}(a), critic-free methods such as GRPO, DAPO, and GSPO avoid training a value model and instead construct response-level advantages~\citep{shaoDeepSeekMathPushingLimits2024,yuDAPOOpenSourceLLM2025,zhengGroupSequencePolicy2025}. These methods are simple and effective, but they assign the same trajectory-level signal to all tokens in a response. In contrast, Proximal Policy Optimization (PPO) remains appealing because its learned critic can in principle provide state-dependent, token-level credit assignment~\citep{schulmanProximalPolicyOptimization2017,schulmanHighDimensionalContinuousControl2018}.

The promise of PPO, however, depends on reliable value estimation under sparse terminal rewards. As shown in Figure~\ref{fig:intro_credit_assignment}(b), the standard critic observes only the prompt and a partial response and must predict whether the incomplete trajectory will produce the correct answer. This prediction is difficult: locally plausible steps may still lead to an incorrect result, while unfinished but promising derivations may appear uncertain. Errors in these state values directly distort the direction and magnitude of the token-level advantages, making critic reliability a central challenge for effective PPO training in long-horizon mathematical reasoning.

We observe an asymmetry in existing PPO-based RLVR systems: reference answers determine terminal rewards but remain unavailable to the critic responsible for token-level credit assignment. Existing value-based methods improve the standard critic through pretraining, GAE variants, or system-level stabilization, yet the critic must still evaluate partial solutions without explicit access to the target answer~\citep{yuanPPOCollapseLongCoT2025,yueVAPOEfficientReliable2025,huOpenReasonerZero2025,liuAsymmetricProximalPolicy2026,shanBringingValueModelsBack2026}. This leaves a naturally available source of supervision underused. As illustrated in Figure~\ref{fig:intro_credit_assignment}(c), the reference answer can serve as training-time privileged information for critic-side value calibration.

In this paper, we propose ReDiPPO, a reference-guided, discrepancy-aware PPO framework for mathematical reasoning. During training, ReDiPPO uses a reference-guided critic that conditions on the prompt, partial response, and reference answer to compute the actor advantage. Conditioning on the target answer provides additional information for estimating whether a partial trajectory will reach the correct outcome. The reference answer is provided only to the critic; the policy retains the standard prompt-only interface.

Reference-guided value calibration also provides a useful discrepancy signal for identifying unreliable credit assignment. ReDiPPO retains a standard critic that evaluates the same token states without the reference answer. The difference between the two critics highlights states that are most sensitive to reference information. ReDiPPO normalizes and clips this discrepancy into a token-level weight and uses it to scale the corresponding advantage in the PPO update. This reweighting emphasizes states at which reference-blind credit assignment is more likely to be unreliable.

We evaluate ReDiPPO on six mathematical reasoning benchmarks—AIME 2024, AIME 2025, AIME 2026, HMMT 2025, Minerva Math, and OlympiadBench—using three policy backbones: Qwen3-4B-Instruct-2507, Qwen3-4B-Thinking-2507, and OLMo3-7B-Instruct-DPO. ReDiPPO achieves the highest average accuracy among the evaluated methods across all three backbones, outperforming vanilla PPO by 1.19--2.37 percentage points. Further analyses show that larger reference--standard discrepancy is associated with longer and less accurate responses, while reference conditioning provides its greatest value-estimation gains late in the reasoning trajectory. Our contributions are summarized as follows:
\begin{itemize}
\item We introduce reference-guided value calibration for PPO-based mathematical reasoning, which uses reference answers to construct a more informative critic baseline.
\item We propose discrepancy-aware token reweighting, which converts the disagreement between reference-guided and standard critics into bounded token-level weights for PPO optimization.
\item Across six benchmarks and three policy backbones, ReDiPPO consistently improves vanilla PPO by 1.19--2.37 percentage points. Further analyses associate larger reference--standard discrepancy with more challenging responses and demonstrate that reference conditioning is particularly beneficial late in the reasoning trajectory.
\end{itemize}

\section{Related Work}
\label{sec:related_work}

\paragraph{RLVR for Mathematical Reasoning.}
Reinforcement learning with verifiable rewards (RLVR) has become a leading approach to mathematical reasoning because final-answer correctness provides an objective and scalable training signal~\citep{shaoDeepSeekMathPushingLimits2024,guoDeepSeekR1IncentivizesReasoning2025}. DeepSeekMath introduced critic-free GRPO, while DAPO and Dr.GRPO improve its sampling, clipping, and normalization, and GSPO moves policy ratios to the sequence level~\citep{yuDAPOOpenSourceLLM2025,liuUnderstandingR1Zero2025,zhengGroupSequencePolicy2025}. In parallel, Open-Reasoner-Zero and VAPO show that critic-based PPO remains competitive when value learning and advantage estimation are carefully designed~\citep{huOpenReasonerZero2025,yueVAPOEfficientReliable2025}. ReDiPPO instead retains PPO's critic and conditions it on the reference answer as training-time privileged information, while keeping the policy prompt-only during training and inference.

\paragraph{Token-Level Credit Assignment.}
Assigning a sparse terminal outcome to individual decisions in a long reasoning trace is a central challenge in RLVR. Critic-free methods broadcast response-level advantages across tokens, avoiding value-estimation errors but providing no explicit distinction among states within a trajectory~\citep{shaoDeepSeekMathPushingLimits2024,yuDAPOOpenSourceLLM2025,zhengGroupSequencePolicy2025}. PPO instead derives token-level advantages from a learned value function, yet long horizons and sparse rewards can bias value estimates and weaken reward propagation~\citep{schulmanProximalPolicyOptimization2017,schulmanHighDimensionalContinuousControl2018,yuanPPOCollapseLongCoT2025}. Recent PPO-based reasoning methods therefore improve value pretraining, value calibration, or advantage estimation, including decoupled and length-adaptive GAE~\citep{yuanPPOCollapseLongCoT2025,yueVAPOEfficientReliable2025,huOpenReasonerZero2025}. Other approaches estimate intermediate values through additional continuations, as in VinePPO~\citep{kazemnejadVinePPORefining2025}, or introduce process-level feedback from step annotations, outcome supervision, or implicit rewards~\citep{lightmanLetsVerifyStep2023,wangMathShepherdVerify2024,cuiProcessReinforcementImplicit2025,sun2026ktae,wu-etal-2026-step,zhu2026gagpogeneralizedadvantagegrouped}. ReDiPPO instead retains terminal rewards, using reference-guided value estimation and critic discrepancy for token reweighting without extra process rewards.

\begin{figure*}[t]
    \centering
    \includegraphics[width=\textwidth]{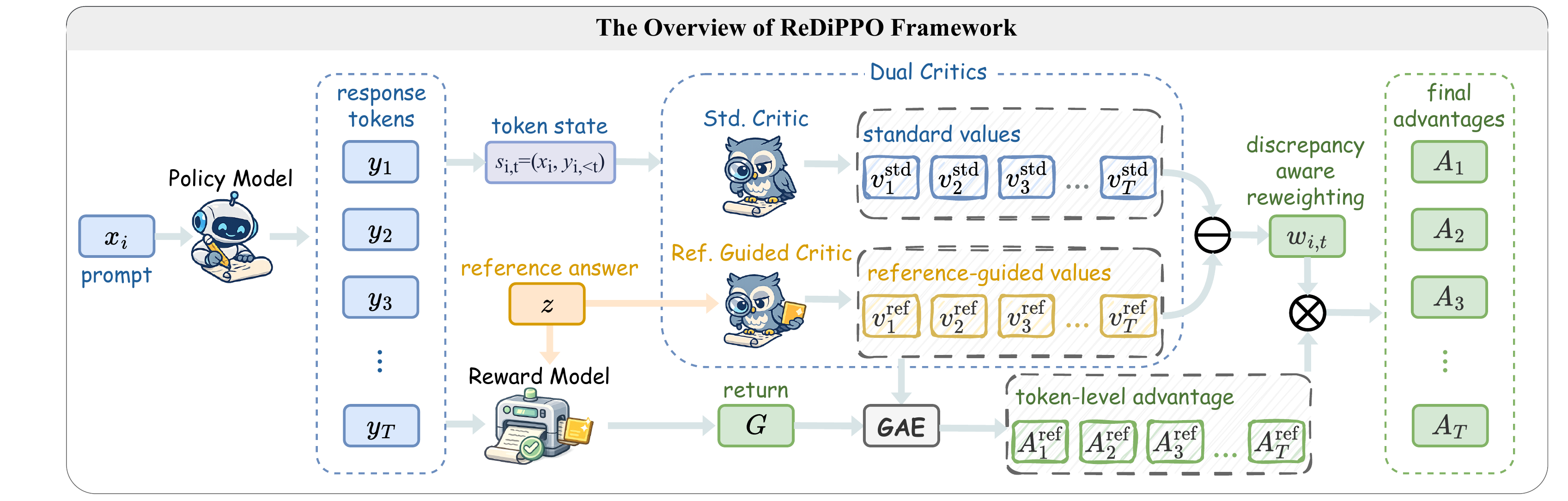}
    \caption{Overview of ReDiPPO. The actor generates responses from the prompt alone, and the verifier provides a terminal outcome. The reference-guided critic uses the reference answer as training-time privileged information to compute the actor advantage, while the standard critic evaluates the same response-token states without the reference. Their token-level discrepancy is normalized and clipped into weights that reweight the reference-guided advantages for PPO optimization.}
    \label{fig:redippo_overview}
\end{figure*}

\section{Method}

\subsection{Preliminaries}
\label{sec:preliminaries}

\paragraph{RLVR for Mathematical Reasoning.}
Each training example consists of a problem prompt and a reference answer $(x,z)\sim\mathcal{D}$.
An autoregressive policy $\pi_\theta$ generates a response $y=(y_1,\ldots,y_T)$ token by token, where the state at step $t$ is the partial solution $s_t=(x,y_{<t})$ and the action is $a_t=y_t$.
After generation, a verifier compares the final answer in $y$ with $z$ and assigns a terminal reward $G=\mathcal{R}(x,y,z)$.
The reinforcement-learning objective is to maximize the expected verifiable reward:
\begin{equation}
    \max_\theta\; J(\theta)
    =
    \mathrm{E}_{(x,z)\sim\mathcal{D},\,y\sim\pi_\theta(\cdot\mid x)}
    \left[\mathcal{R}(x,y,z)\right].
    \label{eq:rlvr_objective}
\end{equation}
Because no direct reward is available for intermediate steps, RLVR must assign this terminal outcome to token decisions made across the full reasoning horizon.

\paragraph{Token-Level PPO.}
PPO~\citep{schulmanProximalPolicyOptimization2017} optimizes the policy with a clipped surrogate objective.
Let $\pi_{\theta_{\mathrm{old}}}$ denote the rollout policy and
$\rho_t(\theta)=\pi_\theta(a_t\mid s_t)/\pi_{\theta_{\mathrm{old}}}(a_t\mid s_t)$ denote the token-level probability ratio.
We write its clipped counterpart as
$\bar{\rho}_t(\theta)=\mathrm{clip}(\rho_t(\theta),1-\epsilon,1+\epsilon)$.
Given an advantage estimate $\hat{A}_t$, PPO maximizes
\begin{equation}
    \mathcal{L}_{\mathrm{PPO}}(\theta)
    =
    \mathrm{E}_{t}\left[
    \min\left(
    \rho_t(\theta)\hat{A}_t,
    \bar{\rho}_t(\theta)\hat{A}_t
    \right)\right].
    \label{eq:ppo_objective}
\end{equation}
The sign of $\hat{A}_t$ determines whether the sampled token should be encouraged or discouraged, while its magnitude controls the token's relative contribution to the update.
Accurate token-level advantage estimation is therefore central to effective PPO optimization.

\paragraph{Critic-Based Advantages.}
Vanilla PPO obtains state-dependent advantages from a critic $V_\phi(s_t)$, commonly through Generalized Advantage Estimation (GAE)~\citep{schulmanHighDimensionalContinuousControl2018}.
GAE accumulates discounted temporal-difference residuals:
\begin{equation}
    \begin{array}{rcl}
        \displaystyle \hat{A}^{\mathrm{GAE}}_t
        & = &
        \displaystyle \sum_{l=0}^{T-t}(\gamma\lambda)^l\delta_{t+l}, \\[0.3em]
        \delta_t
        & = &
        r_t+\gamma V_\phi(s_{t+1})-V_\phi(s_t),
    \end{array}
    \label{eq:gae}
\end{equation}
where $\gamma$ is the reward discount factor and $\lambda$ controls the bias--variance trade-off.
In terminal-only RLVR, $r_t=0$ before completion and the final reward is $G_i$.
Although all tokens in response $y_i$ share the same outcome $G_i$, the critic can assign different advantages to different partial reasoning states.
This enables token-level credit assignment, but an inaccurate value estimate can distort both the direction and strength of the corresponding policy update.

\paragraph{Critic-Free GRPO.}
GRPO~\citep{shaoDeepSeekMathPushingLimits2024,guoDeepSeekR1IncentivizesReasoning2025} removes the critic while retaining a PPO-style clipped policy objective.
For $K$ responses sampled from the same prompt, it constructs
$\hat{A}^{\mathrm{GRPO}}_i=(G_i-\bar{G}_x)/(\sigma_x+\epsilon_{\mathrm{std}})$,
where $\bar{G}_x$ and $\sigma_x$ are the mean and standard deviation of the group returns, and broadcasts this response-level advantage to every valid token in $y_i$.
Removing the critic avoids critic-induced estimation errors but also removes state-dependent distinctions among tokens within the same response.
This exposes the central trade-off addressed by ReDiPPO: PPO supports fine-grained credit assignment but depends critically on reliable value estimation, whereas critic-free methods provide coarser response-level supervision.

\subsection{Overview of ReDiPPO}
\label{sec:method_overview}

ReDiPPO uses reference answers as critic-side privileged information to improve token-level credit assignment while keeping the policy prompt-only during both training and inference. As illustrated in Figure~\ref{fig:redippo_overview}, ReDiPPO retains PPO's clipped actor update and introduces two critics with distinct roles. The reference-guided critic $V_\psi^{\mathrm{ref}}(s_{i,t},z_i)$ supplies the baseline for actor advantages, whereas the standard critic $V_\phi^{\mathrm{std}}(s_{i,t})$ provides a reference-blind comparison value. Their token-level discrepancy is converted into a bounded weight that reweights the reference-guided advantages. Section~\ref{sec:reference_conditioned_value} defines the reference-guided baseline, Section~\ref{sec:discrepancy_token_reweighting} describes discrepancy-aware token reweighting, and Section~\ref{sec:training_pipeline} presents the final actor objective and complete optimization procedure. Algorithm~\ref{alg:redippo} summarizes the complete ReDiPPO optimization procedure.

\subsection{Reference-Guided Value Calibration}
\label{sec:reference_conditioned_value}

\paragraph{Motivation.}
A standard critic must infer the eventual correctness of an incomplete solution from the prompt-response context alone.
Under sparse terminal rewards, locally plausible steps may still lead to an incorrect answer, while an unfinished but promising derivation may appear uncertain.
The reference answer provides a direct target against which the critic can assess whether the evolving trajectory remains compatible with a correct solution.
ReDiPPO therefore uses this information to construct a more informative critic-side baseline without changing the actor input.

\paragraph{Dual-Critic Inputs.}
For each rollout $\tau_i=(x_i,z_i,y_i,G_i)$, both critics evaluate the same valid response-token positions.
The standard critic receives the original context and predicts
\begin{equation}
    v_{i,t}^{\mathrm{std}}
    =
    V_\phi^{\mathrm{std}}(s_{i,t}),
    \qquad
    s_{i,t}=(x_i,y_{i,<t}).
    \label{eq:standard_value}
\end{equation}
The reference-guided critic receives an answer-augmented context, in which $z_i$ is inserted using the template \texttt{The ground truth answer is \{answer\}.}, and predicts
\begin{equation}
    v_{i,t}^{\mathrm{ref}}
    =
    V_\psi^{\mathrm{ref}}(s_{i,t},z_i).
    \label{eq:reference_value}
\end{equation}
Only $v_{i,t}^{\mathrm{ref}}$ is used as the actor baseline; $v_{i,t}^{\mathrm{std}}$ is retained as the reference-blind comparison estimate for discrepancy-aware reweighting.

\paragraph{Dual-Critic Value Learning.}
Both critics are supervised by the same verifier return $G_i$, so their difference arises from the information available to each branch rather than from different regression targets.
For branch $b\in\{\mathrm{std},\mathrm{ref}\}$, let $v_{i,t}^{b,\mathrm{old}}$ denote the value prediction stored before the critic update.
With value clip range $\epsilon_v$, the clipped prediction is
\begin{equation}
    \bar{v}_{i,t}^{b}
    =
    \mathrm{clip}\left(
    v_{i,t}^{b},
    v_{i,t}^{b,\mathrm{old}}-\epsilon_v,
    v_{i,t}^{b,\mathrm{old}}+\epsilon_v
    \right).
    \label{eq:clipped_value_prediction}
\end{equation}
Each branch is optimized with the PPO clipped value loss
\begin{equation}
    \mathcal{L}_{V}^{b}
    =
    \frac{1}{2}\,\mathrm{Agg}_{m}\left[
    \max\left(
    (v_{i,t}^{b}-G_i)^2,
    (\bar{v}_{i,t}^{b}-G_i)^2
    \right)
    \right],
    \label{eq:value_loss}
\end{equation}
where $\mathrm{Agg}_{m}$ denotes masked aggregation over valid response tokens.

\paragraph{Reference-Guided Advantage.}
Because rewards are terminal-only and $\gamma=\lambda=1$, the reference-guided baseline yields
\begin{equation}
    \hat{A}^{\mathrm{ref}}_{i,t}
    =
    G_i-v_{i,t}^{\mathrm{ref}}.
    \label{eq:reference_advantage}
\end{equation}
This advantage retains PPO's state-dependent token-level structure while allowing the critic to judge a partial trajectory against the target answer.
Here, reference-guided value calibration refers to enriching the critic-side information used for the actor baseline; it does not modify the verifier reward.
The reference answer is never appended to the policy input, preventing privileged information from changing the rollout or inference interface.

\subsection{Discrepancy-Aware Token Reweighting}
\label{sec:discrepancy_token_reweighting}

\paragraph{Reference-Standard Discrepancy.}
The standard and reference-guided critics provide two value judgments for the same partial reasoning state under different information conditions.
Let $\mathcal{M}=\{(i,t):m_{i,t}=1\}$ denote all valid response-token positions in a rollout batch.
For each $(i,t)\in\mathcal{M}$, ReDiPPO computes
\begin{equation}
    e_{i,t}
    =
    \left|
    v_{i,t}^{\mathrm{ref}}-v_{i,t}^{\mathrm{std}}
    \right|.
    \label{eq:reference_standard_discrepancy}
\end{equation}
A large $e_{i,t}$ means that access to the reference answer substantially changes the value assigned to that token state.
We therefore use this disagreement as a proxy for states that are difficult for the reference-blind critic to evaluate. Note that it does not directly measure the true critic error, which is unavailable during training.

\paragraph{Normalization and Clipping.}
Raw discrepancies can vary across batches and training stages, so ReDiPPO normalizes them over valid token positions:
\begin{equation}
    \begin{array}{rcl}
        \displaystyle \mu_e
        & = &
        \displaystyle \frac{1}{|\mathcal{M}|}
        \sum_{(i,t)\in\mathcal{M}}e_{i,t}, \\[0.3em]
        \displaystyle \sigma_e^2
        & = &
        \displaystyle \frac{1}{|\mathcal{M}|}
        \sum_{(i,t)\in\mathcal{M}}(e_{i,t}-\mu_e)^2.
    \end{array}
    \label{eq:discrepancy_stats}
\end{equation}
The normalized discrepancy is converted into a weight centered at one:
\begin{equation}
    \omega_{i,t}
    =
    1+
    \frac{e_{i,t}-\mu_e}{\sigma_e+\epsilon_e},
    \label{eq:raw_discrepancy_weight}
\end{equation}
where $\epsilon_e$ prevents numerical instability.
The final positive weight is
\begin{equation}
    w_{i,t}
    =
    \mathrm{clip}(\omega_{i,t},w_{\min},w_{\max}).
    \label{eq:clipped_discrepancy_weight}
\end{equation}
We choose $0<w_{\min}\leq 1\leq w_{\max}$; centering at one preserves ordinary reference-guided PPO as the nominal update, while clipping prevents high-discrepancy positions from dominating optimization.

\begin{algorithm}[tb]
	\small
	\caption{ReDiPPO Training Pipeline}
	\label{alg:redippo}
	\begin{algorithmic}[1]
		\REQUIRE Dataset $\mathcal{D}$, policy $\pi_\theta$, critics $V_\phi^{\mathrm{std}}$ and $V_\psi^{\mathrm{ref}}$, verifier $\mathcal{R}$
		\FOR{each training iteration}
		\STATE Sample $\{(x_i,z_i)\}_{i=1}^{B}\sim\mathcal{D}$; generate $y_i\sim\pi_{\theta_{\mathrm{old}}}(\cdot\mid x_i)$ and obtain $G_i=\mathcal{R}(x_i,y_i,z_i)$.
		\STATE Evaluate $v_{i,t}^{\mathrm{std}}$ and $v_{i,t}^{\mathrm{ref}}$ on aligned valid response-token states using Eqs.~\ref{eq:standard_value}--\ref{eq:reference_value}.
		\STATE Compute $\hat{A}^{\mathrm{ref}}_{i,t}$ with Eq.~\ref{eq:reference_advantage}.
		\STATE Compute $e_{i,t}$ and $w_{i,t}$ using Eqs.~\ref{eq:reference_standard_discrepancy}--\ref{eq:clipped_discrepancy_weight}.
		\STATE Form $\tilde{A}_{i,t}$ with Eq.~\ref{eq:weighted_advantage}.
		\STATE Update both critics using Eq.~\ref{eq:value_loss}.
		\STATE Update $\pi_\theta$ using Eq.~\ref{eq:redippo_actor_objective}.
		\STATE Set $\theta_{\mathrm{old}}\leftarrow\theta$ for the next rollout iteration.
		\ENDFOR
	\end{algorithmic}
\end{algorithm}

\paragraph{Weighted Advantages.}
The discrepancy weight is applied to the reference-guided advantage at the same token position.
We multiply first and then perform the standard PPO masked whitening:
\begin{equation}
    \tilde{A}_{i,t}
    =
    \mathrm{Whiten}_{m}\left(
    w_{i,t}\hat{A}^{\mathrm{ref}}_{i,t}
    \right).
    \label{eq:weighted_advantage}
\end{equation}
Here, $\mathrm{Whiten}_{m}$ subtracts the masked mean and divides by the masked standard deviation over valid response tokens.
The critic predictions, discrepancy weights, and resulting advantages are detached during actor optimization, so the actor update does not backpropagate through either critic.
Consequently, discrepancy-aware reweighting changes the relative contribution of token-level policy-gradient terms while retaining PPO's standard normalized update form.

\subsection{Optimization}
\label{sec:training_pipeline}

ReDiPPO optimizes the PPO clipped surrogate using the weighted advantage $\tilde{A}_{i,t}$:
\begin{equation}
    \mathcal{L}_{\mathrm{ReDiPPO}}(\theta)
    =
    \mathrm{E}_{(i,t)\in\mathcal{M}}\left[
    \min\left(
    \rho_{i,t}\tilde{A}_{i,t},
    \bar{\rho}_{i,t}\tilde{A}_{i,t}
    \right)
    \right].
    \label{eq:redippo_actor_objective}
\end{equation}
Here, $\rho_{i,t}=\pi_\theta(y_{i,t}\mid s_{i,t})/
\pi_{\theta_{\mathrm{old}}}(y_{i,t}\mid s_{i,t})$ and
$\bar{\rho}_{i,t}=\mathrm{clip}(\rho_{i,t},1-\epsilon,1+\epsilon)$.

\section{Experiments}
\label{sec:experiments}

\subsection{Experimental Setup}
\label{sec:experimental_setup}

\paragraph{Training.}
We conduct experiments on three policy backbones spanning instruction-tuned and reasoning-oriented models: \textbf{Qwen3-4B-Instruct-2507}, \textbf{Qwen3-4B-Thinking-2507}~\citep{yang2025qwen3technicalreport}, and \textbf{OLMo3-7B-Instruct-DPO}~\citep{olmo2026olmo3}. All RL methods are trained on mathematical reasoning prompts from DAPO-17K~\citep{yuDAPOOpenSourceLLM2025} and an integer-answer subset of DeepMath-103K~\citep{he2025deepmath103klargescalechallengingdecontaminated}. The latter contains 40,188 problems whose integer final answers support reliable rule-based verification with binary rewards.
For critic-based methods, the critics are initialized from their corresponding policy checkpoints and pretrained for two epochs using the value-pretraining protocol of VAPO~\citep{yueVAPOEfficientReliable2025}. Unless otherwise stated, all methods use the same training data, global batch size of 512, and eight rollout responses per prompt. Training is implemented with \texttt{VeRL}~\citep{sheng2025hybridflow}. We use an actor learning rate of $1\times10^{-6}$ and a critic learning rate of $5\times10^{-6}$. We set $\gamma=\lambda=1$ because rewards are provided only at the end of each response.
Both PPO and ReDiPPO use asymmetric Clip-Higher with $\epsilon_{\mathrm{low}}=0.2$ and $\epsilon_{\mathrm{high}}=0.28$~\citep{yuDAPOOpenSourceLLM2025}, with the KL and entropy-loss coefficients set to zero. ReDiPPO clips discrepancy weights to $[0.5,2.0]$. During rollout, we use a sampling temperature of $1.0$ and maximum response lengths of 8192 tokens for the instruction-tuned models and 32768 tokens for Qwen3-4B-Thinking-2507. Additional critic-pretraining details, baseline configurations, and resource consumption are provided in the supplementary material.

\paragraph{Evaluation.}
We evaluate reasoning performance on AIME 2024, AIME 2025, AIME 2026, HMMT 2025, Minerva Math~\citep{lewkowyczSolvingQuantitative2022}, and OlympiadBench~\citep{heOlympiadBench2024} using the same answer-extraction and verification protocol for all methods. We report avg@32 for AIME and HMMT and avg@8 for the remaining benchmarks.

We evaluate critic quality from two complementary perspectives: value calibration and path selection. First, value explained variance (EV) measures how much of the return variation is explained by the critic. Over valid response-token positions $\mathcal{M}$, it is defined as
\[
\mathrm{EV}
=
1-
\frac{
\mathrm{Var}_{(i,t)\in\mathcal{M}}
\left(R_{i,t}-V_{i,t}\right)
}{
\mathrm{Var}_{(i,t)\in\mathcal{M}}
\left(R_{i,t}\right)+\epsilon
},
\]
where $V_{i,t}$ and $R_{i,t}$ are the predicted value and return target, respectively, and $\epsilon$ ensures numerical stability. A higher EV indicates more accurate value estimation.
Second, path-selection accuracy (PSA) measures whether the critic can identify a correct response among multiple responses to the same prompt. For response $i$, we define its score as the mean value over valid token positions,
\[
s_i
=
\frac{1}{|\mathcal{T}_i|}
\sum_{t\in\mathcal{T}_i}V_{i,t}.
\]
Let $\mathcal{G}_{\mathrm{mix}}$ denote prompt groups containing both correct and incorrect responses. PSA is defined as
\[
\mathrm{PSA}
=
\frac{1}{|\mathcal{G}_{\mathrm{mix}}|}
\sum_{g\in\mathcal{G}_{\mathrm{mix}}}
\mathbf{1}\!\left[G_{\hat{i}_g}=1\right],
\qquad
\hat{i}_g
=
\arg\max_{i\in g}s_i.
\]
Thus, PSA reports how often the response ranked highest by the critic is correct. We restrict this metric to mixed-outcome groups because all-correct and all-incorrect groups do not reveal the critic's ability to distinguish successful reasoning paths. Segment-level EV and PSA further examine how critic reliability changes across different stages of a response; their definitions are provided in the supplementary material.

\paragraph{Baselines.}
We compare ReDiPPO with three types of reference methods. First, we report the initial policy checkpoint before RL training to quantify the overall improvement introduced by RLVR. Second, we include DAPO~\citep{yuDAPOOpenSourceLLM2025} and GSPO~\citep{zhengGroupSequencePolicy2025} as representative critic-free RLVR baselines. Third, we compare with vanilla PPO, which uses the same value-pretraining protocol, training data, rollout budget, and evaluation procedure as ReDiPPO. This matched PPO baseline provides a controlled comparison for assessing the gains associated with reference-guided value calibration and discrepancy-aware token reweighting. Detailed baseline configurations are provided in the supplementary material.

\begin{table*}[!t]
	\centering
	\begin{tabular}{lccccccc}
		\toprule
		\textbf{Model} & \textbf{AIME24} & \textbf{AIME25} & \textbf{AIME26} & \textbf{HMMT25} & \textbf{Minerva} & \textbf{Olympiad} & \textbf{Average} \\
		\midrule
		\textbf{Qwen3-4B-Instruct-2507} & & & & & & & \\
		\quad Vanilla & 54.64 & 41.93 & 49.38 & 27.13 & 60.43 & 72.41 & 50.99 \\
		\quad DAPO & 55.42 & 46.25 & 54.95 & \underline{29.64} & 61.70 & 75.07 & 53.84 \\
		\quad GSPO & \underline{57.76} & \underline{47.60} & \underline{55.99} & 28.12 & \underline{62.18} & \underline{75.10} & \underline{54.46} \\
		\quad PPO & 56.35 & 46.25 & 54.69 & 28.33 & 61.60 & 74.73 & 53.66 \\
		\quad \textbf{ReDiPPO} & \textbf{60.68} & \textbf{49.79} & \textbf{56.77} & \textbf{30.63} & \textbf{62.84} & \textbf{75.45} & \textbf{56.03} \\
		\midrule
		\textbf{Qwen3-4B-Thinking-2507} & & & & & & & \\
		\quad Vanilla & 78.44 & 74.06 & 74.06 & \underline{52.50} & 68.24 & 81.57 & 71.48 \\
		\quad DAPO & 79.69 & \textbf{76.88} & \underline{78.18} & 50.10 & \underline{68.75} & 84.25 & 72.98 \\
		\quad GSPO & \underline{79.90} & 76.35 & 77.19 & 52.24 & 68.52 & \underline{84.72} & \underline{73.15} \\
		\quad PPO & 78.96 & 75.94 & 78.02 & 50.83 & 68.06 & 84.46 & 72.71 \\
		\quad \textbf{ReDiPPO} & \textbf{80.62} & \underline{76.77} & \textbf{78.23} & \textbf{53.33} & \textbf{69.28} & \textbf{85.16} & \textbf{73.90} \\
		\midrule
		\textbf{OLMo3-7B-Instruct-DPO} & & & & & & & \\
		\quad Vanilla & 25.16 & 21.51 & 21.46 & 10.62 & 46.55 & 56.58 & 30.31 \\
		\quad DAPO & \underline{40.89} & \underline{31.82} & \underline{36.77} & \underline{16.88} & \textbf{51.36} & \underline{65.08} & \underline{40.47} \\
		\quad GSPO & 36.82 & 29.79 & 33.80 & 16.56 & 50.83 & 64.09 & 38.65 \\
		\quad PPO & 39.17 & 31.41 & 35.05 & 15.78 & 50.64 & 65.00 & 39.51 \\
		\quad \textbf{ReDiPPO} & \textbf{41.88} & \textbf{33.54} & \textbf{37.24} & \textbf{17.08} & \underline{51.08} & \textbf{65.95} & \textbf{41.13} \\
		\bottomrule
	\end{tabular}
	\caption{Main results on mathematical reasoning benchmarks. We report avg@32 for AIME24, AIME25, AIME26 and HMMT25 and avg@8 for others. All results are presented as percentages. Best results are in bold and second-best results are underlined.}
	\label{tab:main_results}
\end{table*}

\begin{figure*}[!t]
	\centering
	\includegraphics[width=\textwidth]{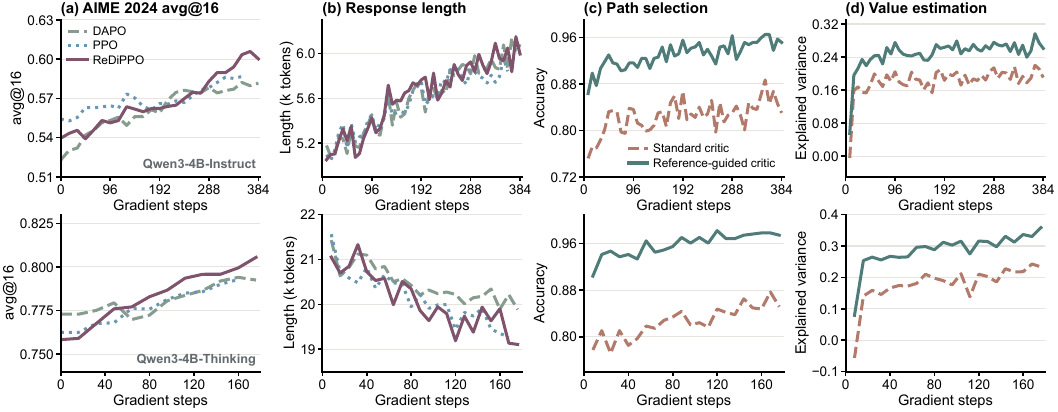}
	\caption{Training dynamics on Qwen3-4B-Instruct (top) and Qwen3-4B-Thinking (bottom). Columns report (a) AIME 2024 avg@16, (b) average response length, (c) training-set path-selection accuracy, and (d) training-set value explained variance. The reference-guided critic achieves higher path-selection accuracy and explained variance than the standard critic.}	\label{fig:training_dynamics}
\end{figure*}

\subsection{Main Results}
\label{sec:main_results}

\paragraph{Overall Performance.}
Table~\ref{tab:main_results} shows that ReDiPPO achieves the highest average score on all three backbones. On Qwen3-4B-Instruct, it ranks first on every benchmark and improves the strongest baseline average from 54.46 to 56.03. On Qwen3-4B-Thinking, ReDiPPO raises the best baseline average from 73.15 to 73.90 and leads on five of the six benchmarks, with DAPO slightly ahead on AIME 2025. ReDiPPO also improves the strongest OLMo3-7B-Instruct-DPO baseline from 40.47 to 41.13 and obtains the best result on five of the six benchmarks. These consistent average gains across instruction-tuned and reasoning-oriented backbones demonstrate the effectiveness of ReDiPPO relative to both critic-free RLVR methods and vanilla PPO.

\paragraph{Training Dynamics.}
Figure~\ref{fig:training_dynamics} compares policy performance, response length, and critic quality on the two Qwen3 backbones. ReDiPPO's response-length trajectory remains close to those of PPO and DAPO on both backbones; on Qwen3-4B-Thinking, its responses become shorter than those of DAPO toward the end of training. At every shared logged step, the reference-guided critic achieves higher training-set path-selection accuracy and explained variance than the standard critic on both backbones. These results show that reference conditioning consistently improves the reported critic-quality metrics while ReDiPPO maintains response lengths in the same range as the compared methods.

\begin{figure}[!t]
\centering
\includegraphics[width=\columnwidth]{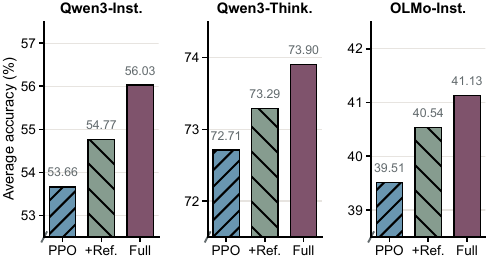}
\caption{Component ablation of ReDiPPO. We report average accuracy across the six benchmarks in Table~\ref{tab:main_results} for PPO, reference-guided PPO (\textsc{+Ref.}), and full ReDiPPO.}
\label{fig:component_ablation}
\end{figure}

\begin{figure}[!t]
\centering
\includegraphics[width=\columnwidth]{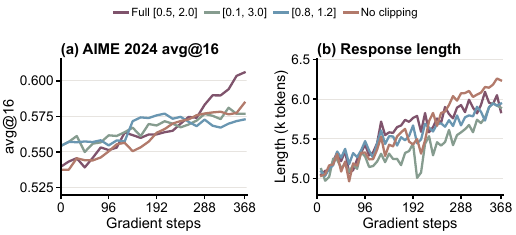}
\caption{Clipping ablation on Qwen3-4B-Instruct. We report (a) AIME 2024 avg@16 and (b) average response length. Full ReDiPPO uses $[w_{\min},w_{\max}]=[0.5,2.0]$; the alternatives use $[0.1,3.0]$, $[0.8,1.2]$, or no clipping.}
\label{fig:clipping_ablation}
\end{figure}

\subsection{Ablation Studies}
\label{sec:ablation_studies}

\paragraph{Component Contributions.}
Figure~\ref{fig:component_ablation} shows that reference-guided value calibration consistently improves PPO, yielding gains of 1.11, 0.58, and 1.03 average-accuracy points on Qwen3-Inst., Qwen3-Think., and OLMo-Inst., respectively. Discrepancy-aware token reweighting provides a further 1.26, 0.61, and 0.59 points, increasing the total gains over PPO to 2.37, 1.19, and 1.62 points. The consistent incremental improvements across all three backbones indicate that value calibration and token reweighting make complementary contributions to ReDiPPO.

\paragraph{Effect of Weight Clipping.}
Figure~\ref{fig:clipping_ablation} examines the sensitivity of ReDiPPO to the discrepancy-weight bounds. At the shared training horizon, the default range $[0.5,2.0]$ achieves the highest smoothed AIME 2024 avg@16 while maintaining a shorter average response length than the unclipped variant. Both widening the range to $[0.1,3.0]$ and tightening it to $[0.8,1.2]$ yield lower avg@16, whereas removing clipping produces the longest responses without matching the accuracy of the default configuration. These results show that, across the tested configurations, the clipping bounds materially affect training behavior and that the selected range provides the strongest observed accuracy while avoiding the response-length growth seen without clipping.

\subsection{Analysis of Reference-Standard Discrepancy}
\label{sec:discrepancy_analysis}

We analyze reference-standard discrepancy at both the response and token levels to determine what it captures and where it is most informative.

\begin{figure}[!t]
\centering
\includegraphics[width=\columnwidth]{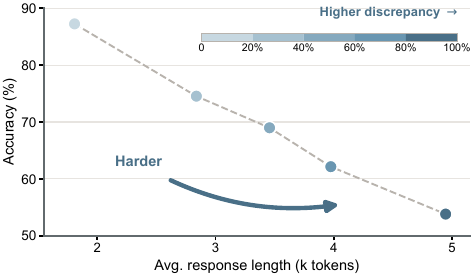}
\caption{Response-level difficulty across quintiles of mean reference-standard discrepancy on Qwen3-4B-Instruct. Each point reports accuracy and length over responses from the test sets in Table~\ref{tab:main_results}; darker colors denote higher discrepancy. Higher-discrepancy groups are longer and less accurate.}
\label{fig:discrepancy_quintiles}
\end{figure}

\paragraph{Response-Level Difficulty.}
Figure~\ref{fig:discrepancy_quintiles} groups Qwen3-4B-Instruct responses into quintiles by their mean token-level discrepancy. From the lowest to the highest quintile, average response length increases from 1,808 to 4,946 tokens ($2.74\times$), while accuracy decreases from 87.25\% to 53.81\%. Longer traces and lower success rates jointly characterize these responses as more difficult. This monotonic pattern identifies reference-standard discrepancy as an empirical indicator of challenging reasoning trajectories.

\begin{figure}[t]
	\centering
	\includegraphics[width=\columnwidth]{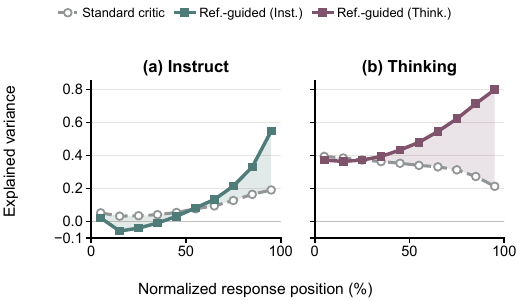}
	\caption{Position-wise critic explained variance on the six test sets in Table~\ref{tab:main_results}. Responses are divided into ten equal-length bins. The reference-guided critic shows increasingly larger gains over the standard critic toward the end of the response on both Qwen3-Instruct and Qwen3-Thinking.}
	\label{fig:discrepancy_analysis}
\end{figure}

\paragraph{Token-Level Localization.}
Figure~\ref{fig:discrepancy_analysis} localizes the benefit of reference conditioning along the response. The reference-guided critic gains its largest advantage in later positions: in the final bin, its EV reaches 0.548 versus 0.191 for the standard critic on Qwen3-Instruct, and 0.798 versus 0.213 on Qwen3-Thinking. Because later states contain more evidence about the outcome, this gap is especially informative for credit assignment. It shows where reference information most changes the value assessment and motivates token-level reweighting instead of assigning one difficulty score to the entire response.

\paragraph{Takeaway.}
The response-level analysis identifies which trajectories are more challenging, while the position-wise analysis identifies where reference information is most useful within them. Both findings support using reference-standard discrepancy as a token-level training signal.

\section{Conclusion}

This paper introduced ReDiPPO to address unreliable token-level credit assignment in mathematical RLVR. ReDiPPO uses reference answers as critic-side privileged information to construct the actor baseline and reweights token-level advantages according to the disagreement between reference-guided and standard critics, while keeping the policy prompt-only. Across six benchmarks and three model backbones, ReDiPPO achieves the highest average accuracy among the evaluated methods and improves vanilla PPO by 1.19--2.37 percentage points. Ablations confirm consistent contributions from both components, while further analyses link larger discrepancy to more challenging responses and stronger late-stage gains from reference conditioning.
An open question is whether reference-guided value learning transfers to domains without unique, rule-verifiable answers. We hope this work motivates further study of training-time reference information for PPO credit assignment.

\clearpage
\bibliography{references}

\clearpage
\appendix

\section*{Contents}

\begingroup
\renewcommand{\arraystretch}{1.35}
\begin{tabular*}{\columnwidth}{@{}p{0.84\columnwidth}@{\extracolsep{\fill}}r@{}}
\textbf{Appendix~\ref{app:llm_usage}}\quad Usage of LLMs
    & \pageref{app:llm_usage} \\
\textbf{Appendix~\ref{app:critic_pretraining}}\quad Critic Pretraining
    & \pageref{app:critic_pretraining} \\
\textbf{Appendix~\ref{app:experimental_details}}\quad Additional Experimental Details
    & \pageref{app:experimental_details} \\
\hspace*{1.5em}\ref{app:datasets}\quad Datasets
    & \pageref{app:datasets} \\
\hspace*{1.5em}\ref{app:evaluation_details}\quad Evaluation Details
    & \pageref{app:evaluation_details} \\
\hspace*{1.5em}\ref{app:baselines}\quad Baselines
    & \pageref{app:baselines} \\
\hspace*{1.5em}\ref{app:training_details}\quad Training Details
    & \pageref{app:training_details} \\
\textbf{Appendix~\ref{app:supplementary_results}}\quad Supplementary Results
    & \pageref{app:supplementary_results} \\
\hspace*{1.5em}\ref{app:reference_information_ablation}\quad Form of Privileged Reference Information
    & \pageref{app:reference_information_ablation} \\
\hspace*{1.5em}\ref{app:resource_consumption}\quad Analysis of Computational Efficiency
    & \pageref{app:resource_consumption} \\
\end{tabular*}
\endgroup

\section{Usage of LLMs}
\label{app:llm_usage}

During the preparation of this manuscript, large language model (LLM)-based tools were used to assist with language polishing, structural editing, and \LaTeX{} formatting. All LLM-assisted text was reviewed and revised by the authors. The scientific claims, methodological decisions, experimental design, result interpretation, and conclusions were determined and verified by the authors, who take full responsibility for the final manuscript.

\section{Critic Pretraining}
\label{app:critic_pretraining}

\paragraph{Pretraining Data.}
For each policy backbone, we construct its critic-pretraining data by using the corresponding policy checkpoint to sample eight responses for every prompt in the training set. Each response is assigned a binary label by the rule-based verifier described in Appendix~\ref{app:evaluation_details}. Because these policy checkpoints already achieve relatively high accuracy, incorrect responses are underrepresented in the resulting data. We therefore upsample the incorrect response trajectories to obtain an approximately balanced ratio of correct and incorrect examples.

\paragraph{Pretraining Protocol.}
For each policy backbone, the standard and reference-guided critics are initialized from the same checkpoint as the policy. Both critics undergo value pretraining for two epochs before RL training, following VAPO~\citep{yueVAPOEfficientReliable2025}. During value pretraining, both critics minimize the mean squared error (MSE) between their token-level value predictions and the corresponding verifier-return targets. They differ only in their inputs: the standard critic observes the prompt and partial response, whereas the reference-guided critic additionally observes the reference answer. The PPO baseline uses the same critic initialization and value-pretraining protocol as ReDiPPO.

\section{Additional Experimental Details}
\label{app:experimental_details}

\subsection{Datasets}
\label{app:datasets}

We train all RL methods on DAPO-17K~\citep{yuDAPOOpenSourceLLM2025} and an integer-answer subset of DeepMath-103K~\citep{he2025deepmath103klargescalechallengingdecontaminated}. The DeepMath subset contains 40,188 mathematical reasoning problems whose final answers are integers. Each training example provides a problem prompt and a reference answer. Before RL training, we filter the prompt pool using the eight responses sampled by the corresponding policy checkpoint during critic-data construction. We discard prompts for which all eight responses are correct, while retaining both all-incorrect prompts and prompts with mixed outcomes. This preprocessing removes problems that the initial policy has already mastered while preserving unsolved and partially solved problems for RL. For evaluation, we use AIME 2024, AIME 2025, AIME 2026, HMMT 2025, Minerva Math, and OlympiadBench. Table~\ref{tab:dataset_statistics} lists the datasets and models used in our experiments, together with their URLs, licenses, and dataset sizes.

\begin{table*}[!t]
\centering
\caption{Dataset and model assets used in our experiments. Licenses correspond to the linked distributions.}
\label{tab:dataset_statistics}
\footnotesize
\setlength{\tabcolsep}{3.5pt}
\renewcommand{\arraystretch}{1.1}
\begin{tabular}{@{}lp{0.20\textwidth}p{0.42\textwidth}p{0.17\textwidth}r@{}}
\toprule
\textbf{Category} & \textbf{Asset} & \textbf{URL} & \textbf{License Name} & \textbf{Questions} \\
\midrule
Dataset & DAPO-17K & \url{https://huggingface.co/datasets/BytedTsinghua-SIA/DAPO-Math-17k} & Apache-2.0 & 17,398 \\
Dataset & DeepMath-103K (subset) & \url{https://huggingface.co/datasets/zwhe99/DeepMath-103K} & MIT & 40,188 \\
Dataset & AIME 2024 & \url{https://huggingface.co/datasets/math-ai/aime24} & Apache-2.0 & 30 \\
Dataset & AIME 2025 & \url{https://huggingface.co/datasets/math-ai/aime25} & Apache-2.0 & 30 \\
Dataset & AIME 2026 & \url{https://huggingface.co/datasets/math-ai/aime26} & Apache-2.0 & 30 \\
Dataset & HMMT 2025 & \url{https://huggingface.co/datasets/FlagEval/HMMT_2025} & CC BY-NC-SA 4.0 & 30 \\
Dataset & Minerva Math & \url{https://huggingface.co/datasets/math-ai/minervamath} & MIT & 272 \\
Dataset & OlympiadBench & \url{https://huggingface.co/datasets/math-ai/olympiadbench} & MIT & 675 \\
\midrule
Model & Qwen3-4B-Instruct-2507 & \url{https://huggingface.co/Qwen/Qwen3-4B-Instruct-2507} & Apache-2.0 & -- \\
Model & Qwen3-4B-Thinking-2507 & \url{https://huggingface.co/Qwen/Qwen3-4B-Thinking-2507} & Apache-2.0 & -- \\
Model & OLMo3-7B-Instruct-DPO & \url{https://huggingface.co/allenai/Olmo-3-7B-Instruct-DPO} & Apache-2.0 & -- \\
Model & CompassVerifier-3B & \url{https://huggingface.co/opencompass/CompassVerifier-3B} & Apache-2.0 & -- \\
\bottomrule
\end{tabular}
\end{table*}

\begin{figure*}[t]
\centering
\includegraphics[width=\textwidth]{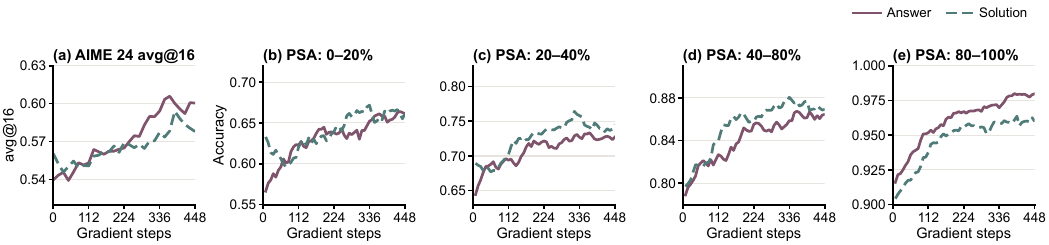}
\caption{Ablation on the form of privileged reference information. We compare the concise reference answer with a verifier-correct reference solution. Panel (a) reports AIME 2024 avg@16 with an exponential moving average (EMA; $\alpha=0.35$); panels (b)--(e) report training-set PSA over four response-position segments with EMA smoothing ($\alpha=0.20$). All panels use the common training horizon of 448 gradient steps.}
\label{fig:reference_information_ablation}
\end{figure*}

\subsection{Evaluation Details}
\label{app:evaluation_details}

\paragraph{Training Stage.}
Because the training data contain only integer reference answers, we use a rule-based verifier during training. It extracts the answer enclosed in the final \texttt{\textbackslash boxed\{...\}} expression of each generated response and matches it against the reference integer, producing a binary outcome reward in $\{0,1\}$.

\paragraph{Test Stage.}
At test time, we use a two-stage verification procedure that combines \texttt{math-verify}\footnote{\url{https://github.com/huggingface/Math-Verify}} with model-based judgment. We first pass every predicted answer through \texttt{math-verify}. Responses judged correct by \texttt{math-verify} are accepted directly, while those judged incorrect are sent to \texttt{CompassVerifier-3B} for a second verification pass. For these responses, the second-stage judgment determines the final correctness label.

\subsection{Baselines}
\label{app:baselines}

We compare ReDiPPO against three groups of baselines. First, we report each vanilla policy checkpoint without additional RL training, which measures the starting capability of each backbone. Second, we include critic-free RLVR methods, including DAPO~\citep{yuDAPOOpenSourceLLM2025} and GSPO~\citep{zhengGroupSequencePolicy2025}, to compare against strong sequence-level optimization approaches that avoid value-model training. Third, we include vanilla PPO with the same critic pretraining protocol, training data, rollout budget, and evaluation protocol as ReDiPPO. Together, these baselines compare ReDiPPO with both the untrained starting point and representative critic-free and critic-based RLVR methods under a consistent experimental protocol.

\subsection{Training Details}
\label{app:training_details}

All RL methods are implemented in \texttt{VeRL}~\citep{sheng2025hybridflow} and trained in an off-policy setting.

\paragraph{Optimization Settings.}
We use a global batch size of 512 and set the actor and critic learning rates to $1\times10^{-6}$ and $5\times10^{-6}$, respectively. We set $\gamma=\lambda=1$ for terminal-reward propagation. PPO and ReDiPPO both use asymmetric Clip-Higher~\citep{yuDAPOOpenSourceLLM2025} with $\epsilon_{\mathrm{low}}=0.2$ and $\epsilon_{\mathrm{high}}=0.28$. The KL and entropy-loss coefficients are set to zero. For ReDiPPO, discrepancy weights are clipped with $w_{\min}=0.5$ and $w_{\max}=2.0$.

\paragraph{Rollout Settings.}
For each prompt, we sample eight responses with a temperature of $1.0$. We enable dynamic sampling during RL: prompt groups whose current eight responses are either all correct or all incorrect are excluded from the update batch, so optimization focuses on prompts that produce both positive and negative learning signals under the current policy. The maximum response length is 8,192 tokens for Qwen3-4B-Instruct and OLMo3-7B-Instruct, and 32,768 tokens for Qwen3-4B-Thinking. All compared methods use the same training data, dynamic-sampling rule, and rollout budget.

\paragraph{Compute Infrastructure.}
We run the Qwen3-4B-Instruct experiments on 8 H200 GPUs, and the Qwen3-4B-Thinking and OLMo3-7B-Instruct experiments on 16 H200 GPUs.

Table~\ref{tab:key_hyperparameters} summarizes the key hyperparameters used in our experiments. Method-specific settings are identified in the corresponding row labels; the remaining settings are shared by PPO and ReDiPPO.

\begin{table}[t]
\centering
\caption{Key hyperparameters used in our experiments.}
\label{tab:key_hyperparameters}
\small
\setlength{\tabcolsep}{5pt}
\renewcommand{\arraystretch}{1.08}
\begin{tabular}{@{}p{0.64\columnwidth}l@{}}
\toprule
\textbf{Parameter} & \textbf{Value} \\
\midrule
Critic pretraining epochs & 2 \\
Global batch size & 512 \\
Actor learning rate & $1\times10^{-6}$ \\
Critic learning rate & $5\times10^{-6}$ \\
Discount factor $\gamma$ & $1.0$ \\
GAE parameter $\lambda$ & $1.0$ \\
Clip low $\epsilon_{\mathrm{low}}$ (DAPO / PPO / ReDiPPO) & $0.2$ \\
Clip high $\epsilon_{\mathrm{high}}$ (DAPO / PPO / ReDiPPO) & $0.28$ \\
Clip low $\epsilon_{\mathrm{low}}$ (GSPO) & $0.0003$ \\
Clip high $\epsilon_{\mathrm{high}}$ (GSPO) & $0.0004$ \\
KL loss coefficient & $0$ \\
Entropy loss coefficient & $0$ \\
Discrepancy weight lower bound $w_{\min}$ & $0.5$ \\
Discrepancy weight upper bound $w_{\max}$ & $2.0$ \\
Responses per prompt & 8 \\
Rollout temperature & $1.0$ \\
Max response length (Instruct) & 8,192 \\
Max response length (Thinking) & 32,768 \\
\bottomrule
\end{tabular}
\end{table}

\section{Supplementary Results}
\label{app:supplementary_results}

\subsection{Form of Privileged Reference Information}
\label{app:reference_information_ablation}

ReDiPPO uses the concise reference answer as privileged critic-side information. On Qwen3-4B-Instruct-2507, we ablate this choice against a richer reference solution while keeping the remaining training configuration unchanged. The answer variant uses the main-paper template \texttt{The ground truth answer is \{answer\}.} For the solution variant, we use Qwen3-235B-A22B-Thinking~\citep{yang2025qwen3technicalreport} to generate a solution before RL training and retain only the final summary after \texttt{\textless/think\textgreater}. We keep summaries whose extracted answers are verifier-correct and whose lengths fall in the inclusive range of $[50,2050]$ tokens. The resulting text is appended to the critic prompt using the exact template \texttt{A reference solution is \{solution\}}. This reference is supplied only to the critic; the policy input remains unchanged.

To localize the effect along a response, we compute segment-level path-selection accuracy (PSA). For a normalized response interval $[a,b)$, let $\mathcal{T}_i^{[a,b)}$ be the valid tokens of response $i$ that fall in that interval. We replace the full-response score in the main-paper PSA definition with
\[
s_i^{[a,b)}
=
\frac{1}{|\mathcal{T}_i^{[a,b)}|}
\sum_{t\in\mathcal{T}_i^{[a,b)}}V_{i,t},
\]
and otherwise use the same mixed-outcome prompt groups and top-scoring-response criterion. Figure~\ref{fig:reference_information_ablation} reports the four intervals used in this ablation.

The richer solution improves critic path selection over the middle of the response but not near its end. Across the 56 shared PSA checkpoints, the solution variant is more accurate at 37 checkpoints in the 20--40\% segment and 42 checkpoints in the 40--80\% segment, compared with 17 and 14 checkpoints for the answer variant, respectively. The first 20\% segment has no consistent winner. In contrast, the answer variant is more accurate at 50 of 56 checkpoints in the final 20\% segment, while the solution variant leads at only five. One possible interpretation of this position-dependent reversal is that detailed derivations help assess intermediate progress, whereas the concise answer provides a cleaner signal for judging whether a nearly complete trajectory reaches the correct result.

Stronger middle-segment PSA does not translate into better downstream policy performance in this setting. Averaged over the last five shared AIME 2024 evaluation checkpoints, the answer and solution variants obtain 59.58 and 58.46 avg@16, respectively, a 1.13-point advantage for the concise answer. The ablation therefore supports using the reference answer as ReDiPPO's default privileged signal: it is cheaper to construct, produces stronger late-response path selection, and yields slightly better downstream accuracy in this run. Because this comparison uses one generated-solution pipeline and a single training run per variant, it does not establish that longer reference information is universally harmful.

One possible explanation for the solution variant's weaker downstream result is that a single precomputed solution does not represent the diversity of valid reasoning paths. Conditioning on one derivation may bias the critic toward that particular path and undervalue different yet correct trajectories. We leave testing this hypothesis and developing path-diverse or path-invariant privileged reference signals to future work.

\subsection{Analysis of Computational Efficiency}
\label{app:resource_consumption}

\begin{figure*}[!t]
\centering
\includegraphics[width=\textwidth]{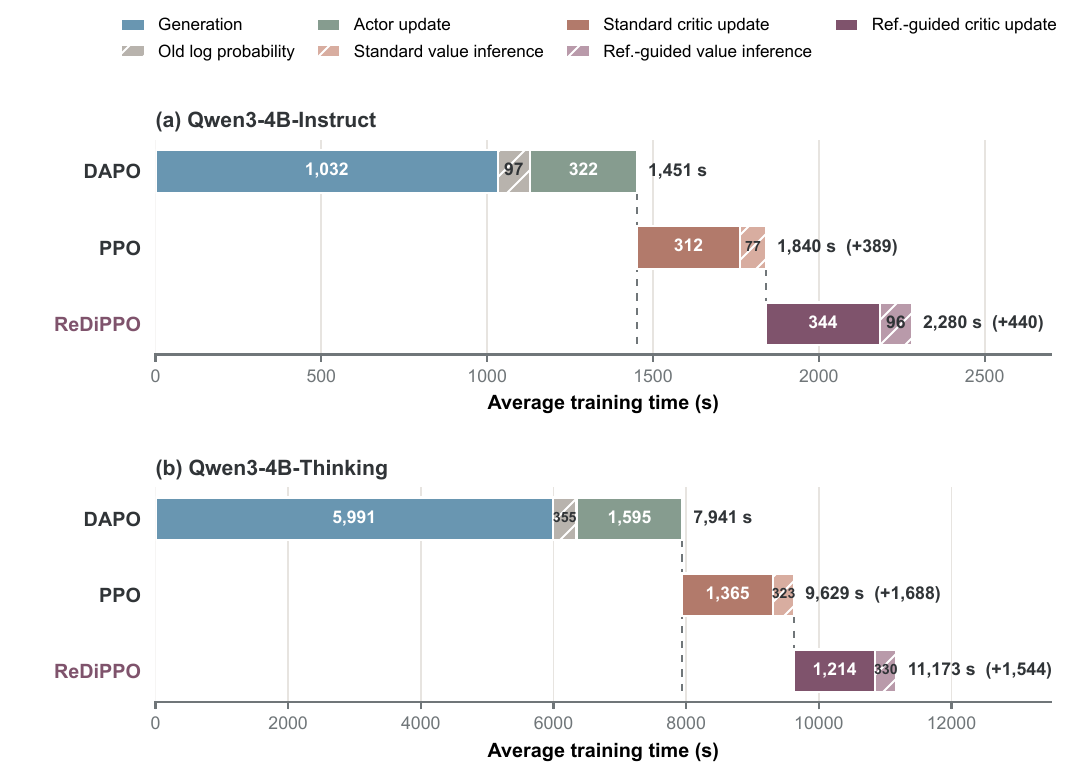}
\caption{Average incremental training-time breakdown of DAPO, PPO, and ReDiPPO on (a) Qwen3-4B-Instruct and (b) Qwen3-4B-Thinking. DAPO includes response generation, old-policy log-probability evaluation, and the actor update. PPO adds a standard-critic update and value inference, while ReDiPPO further adds a reference-guided-critic update and value inference. Dashed lines mark inherited cumulative time; totals and incremental overheads are shown at each stage.}
\label{fig:training_time_breakdown}
\end{figure*}

Figure~\ref{fig:training_time_breakdown} reports the average training-time breakdown for Qwen3-4B-Instruct and Qwen3-4B-Thinking. Response generation is the dominant individual component for both backbones: it takes 1,032 seconds on Qwen3-4B-Instruct and 5,991 seconds on Qwen3-4B-Thinking, accounting for 71.1\% and 75.4\% of their respective DAPO runtimes. Because this actor-side cost is shared by all three methods, generation remains the largest single component of the cumulative PPO and ReDiPPO runtimes, accounting for 56.1\% and 45.3\% on Qwen3-4B-Instruct and 62.2\% and 53.6\% on Qwen3-4B-Thinking, respectively. The two panels use the model-specific compute configurations reported in Appendix~\ref{app:training_details}; the absolute wall-clock times should therefore be compared within each panel rather than across backbones.

On Qwen3-4B-Instruct, PPO adds 312 seconds for the standard-critic update and 77 seconds for value inference, increasing the DAPO time by 389 seconds (26.8\%) to 1,840 seconds. ReDiPPO further adds 344 seconds for the reference-guided-critic update and 96 seconds for its value inference, increasing the PPO time by 440 seconds (23.9\%) to 2,280 seconds. On Qwen3-4B-Thinking, PPO adds 1,365 seconds for the standard-critic update and 323 seconds for value inference, increasing the DAPO time by 1,688 seconds (21.3\%) to 9,629 seconds. ReDiPPO further adds 1,214 seconds for the reference-guided-critic update and 330 seconds for its value inference, increasing the PPO time by 1,544 seconds (16.0\%) to 11,173 seconds. The shared actor stages are drawn only once in each panel, while the dashed guides indicate the cumulative cost inherited by each method. This overhead is confined to training: both critics are discarded at inference, so ReDiPPO introduces no additional critic-side inference computation.

\end{document}